\documentclass[letterpaper]{article}
\usepackage{aaai}
\usepackage{times}
\usepackage{helvet}
\usepackage{courier}
\usepackage{booktabs} 
\pdfoutput=1
\usepackage{times}  
\usepackage{helvet}  
\usepackage{courier}  
\usepackage{graphicx}  
\usepackage{mathtools}
\usepackage{booktabs}
\usepackage{color}
\usepackage{hyperref}
\usepackage{dblfloatfix}
\usepackage{etoolbox}
\usepackage{calc}
\usepackage{xcolor}
\usepackage[ruled, noline, longend, linesnumbered, boxed]{algorithm2e}
\usepackage{mdwlist}
\usepackage[skins]{tcolorbox}
\usepackage{xspace}
\usepackage{latexsym}
\usepackage{amsfonts}
\usepackage{amsmath}
\usepackage{amssymb}
\usepackage{color}
\usepackage{colortbl}
\usepackage{epsfig}
\usepackage{graphicx}
\usepackage{pbox}
\usepackage{array,multirow,graphicx}
\usepackage{paralist}
\usepackage{enumerate}
\usepackage[color,matrix,arrow,all]{xy}
\usepackage{comment}
\usepackage{booktabs}
\usepackage{balance}
\usepackage{stmaryrd}
\usepackage{pifont}
\usepackage{hhline}
\usepackage{listings}
\usepackage{array}
\usepackage{float}
\usepackage[flushleft]{threeparttable}
\usepackage{graphicx,wrapfig,lipsum}
 \usepackage[framemethod=TikZ]{mdframed}
\usepackage{subcaption}
\usetikzlibrary{arrows}

\DeclareMathAlphabet{\pazocal}{OMS}{zplm}{m}{n}

\usepackage{array}
\newcolumntype{M}{>{\begin{varwidth}{2cm}}l<{\end{varwidth}}}
\newcolumntype{L}[1]{>{\raggedright\let\newline\\\arraybackslash\hspace{0pt}}m{#1}}
\newcolumntype{C}[1]{>{\centering\let\newline\\\arraybackslash\hspace{0pt}}m{#1}}
\newcolumntype{R}[1]{>{\raggedleft\let\newline\\\arraybackslash\hspace{0pt}}m{#1}}

\newcommand{\ie}{{\em i.e.,}\xspace}
\newcommand{\eg}{{\em e.g.,}\xspace}

\newenvironment{noindlist}
 {\begin{list}{\labelitemi}{\leftmargin=0em \itemindent=1em}}
 {\end{list}}

\newcommand{\blue}[1]{\textcolor{blue}{#1}}

\newcommand{\Ni}{({\em i})~}
\newcommand{\Nii}{({\em ii})~}
\newcommand{\Niii}{({\em iii})~}

\newcommand{\Niv}{({\em iv})~}

\newcommand{\acttovec}{\emph{activity2vec}}
\newcommand{\sampletovec}{\texttt{sample2vec}}
\newcommand{\hourtovec}{\texttt{hour2vec}}
\newcommand{\daytovec}{\texttt{day2vec}}
\newcommand{\weektovec}{\texttt{week2vec}}

\newcommand{\sampletovecreg}{\texttt{sample2vec+Reg}}
\newcommand{\hourtovecreg}{\texttt{hour2vec+Reg}}
\newcommand{\daytovecreg}{\texttt{day2vec+Reg}}

\DeclareMathOperator \real{\mathbb{R}}

\newcommand{\Ls}{\mathcal{L}}

\newcommand{\Ns}{\pazocal{N}}

\newcommand{\ph}{\mathbf{\Phi}}
\newcommand\norm[1]{\left\lVert#1\right\rVert}

\begin{document}

\title{Adversarial Unsupervised Representation Learning for Activity Time-Series}
\author{Karan Aggarwal\textsuperscript{1}, Shafiq Joty\textsuperscript{2}, Luis Fernandez-Luque\textsuperscript{3}, Jaideep Srivastava\textsuperscript{1}\\
\textsuperscript{1}University of Minnesota, \textsuperscript{2}Nanyang Technological University, \textsuperscript{3}Qatar Computing Research Institute \\
aggar081@umn.edu, srjoty@ntu.edu.sg, lluque@hbku.edu.qa, srivasta@umn.edu\\
}

\maketitle

\begin{abstract}

Sufficient physical activity and restful sleep play a major role in the prevention and cure of many chronic conditions. Being able to proactively screen and monitor such chronic conditions would be a big step forward for overall health. 
The rapid increase in the popularity of wearable devices provides a significant new source,  making it possible to track the user's lifestyle real-time. 
In this paper, we propose a novel unsupervised representation learning technique called \acttovec\ that learns and ``summarizes" the discrete-valued activity time-series.  It learns the representations with three components: \Ni the co-occurrence and magnitude of the activity levels in a time-segment, \Nii neighboring context of {the} time-segment, and \Niii {promoting} subject-invariance with adversarial training. 
 We evaluate our method on four disorder prediction tasks using linear classifiers. Empirical evaluation
demonstrates that our proposed method scales and performs
better than many strong baselines. The adversarial
regime helps improve the generalizability of our representations by promoting subject invariant features. We also show that using the representations at the level of a day works the best since human activity is structured in terms of daily routines. 
 
\end{abstract}

%
%

\section{Introduction}
\label{sec:intro}
\vspace{-0.25em}
Physical activity and sleep are crucial to health and wellbeing. Requisite activity and sufficient sleep prevent various illnesses such as diabetes and depression~\cite{warburton2006health}. Rise in chronic conditions, mainly due to aging and unhealthy lifestyles, is putting our healthcare system under stress.
In the current approach to sleep-disorder screening subjects have to go through different diagnosis steps, involving questionnaires and \emph{polysomnography} (PSG). With increasing popularity of wearable devices like \textit{Fitbit}, which collect detailed data about the body's movements, there is an increased interest in using actigraphy for detecting sleep-related disorders and tracking longitudinal changes in the subject's condition. Although much lower fidelity than clinical devices, availability of wearables provides a novel opportunity,  owing to its non-intrusive and real-time capabilities; specifically, if we can develop techniques to extract information from the vast amounts of body-monitoring data. Such techniques could be useful to assist healthcare professionals as well as help monitor behavioral therapy, e.g., exercise. \emph{This application is an essential motivation of this work.} 

\noindent Only a minuscule proportion of the population has both their clinical data and wearables data available. Thus a purely supervised approach utilizing the wearables-clinical data corpus is sub-optimal since it renders the activity data from a majority of subjects redundant. Hence, any approach towards using activity signals should utilize the unsupervised learning. Second, an important aspect is that information in actigraphy signals depends on the \emph{subjects} and their \emph{environments}, such as their routines and surroundings~\cite{storm2015step} along with measurement errors owing to device design. 
We propose a two-pronged approach. First, a mapping of the temporal relations of the activities from the time-series to a feature space should be learned. Second, this feature space should take into account the subject's environment, and make the representations invariant to the subject and their environment.     

\noindent We propose a new method 
that addresses the challenges mentioned above. The core of our method is an unsupervised representation learning model, \acttovec\ that learns \emph{distributed} representations for activity signals spanning over a time segment (\eg\ at a day level) in a subject invariant manner.  
We use two public data-sets to evaluate our approach against baselines on four disorder prediction tasks. Using a linear classifier (logistic regression), we show that our proposed representation learning method outperforms the baseline time-series representation methods with a good margin, with day-level time-segment representations performing the best. The \textbf{linear} classifier with our learned features \emph{performs at par with the non-linear convolution neural network} baseline trained end-to-end on the classification tasks. We also demonstrate the effectiveness of the subject-invariance loss in inducing subject invariant features. 


Unlike traditional time-series methods, our feature vectors can be used to boost the performance of the supervised learning models. It has been shown that using unsupervised pre-trained vectors to initialize the supervised models produces better performance~\cite{hinton2012deep}. Our method is general enough to be applied to other domains with similar time-series like activity tracking through smart-phones, traffic monitoring, or other types of sensor data.  
Specifically, we make the following contributions:
\begin{noindlist}
\item \textbf{Unsupervised scalable embeddings for activity time-series:} To utilize enormous unlabeled human activity data, we propose a novel unsupervised representation learning method \acttovec\ 
that uncovers activity patterns through \emph{distributed} representation in a scalable fashion, which can be leveraged towards prediction tasks.

\item \textbf{A {hierarchical} model of representation:} One of the persistent challenges in the time-series domain is the selection of time-segments granularity that serve as the primary analysis units. We explore learning representations at various levels of time granularity. We devise a novel algorithm that optimizes two different measures to capture local and global patterns in the activity time-series, along with an ordinal loss to account for the natural ordering in the activity signals.


\item \textbf{Subject-invariant representations:} 
The noise from the subjects environments can hinder the generalization. In order to make the representations invariant to subject environments, we train our representations with a subject invariance loss in the adversarial training setting.



\end{noindlist}

\section{Related Work} \label{sec:rel}
\begin{figure}[t!]
  \centering
      \vspace{-1.5em}
\includegraphics[width=0.45\textwidth]{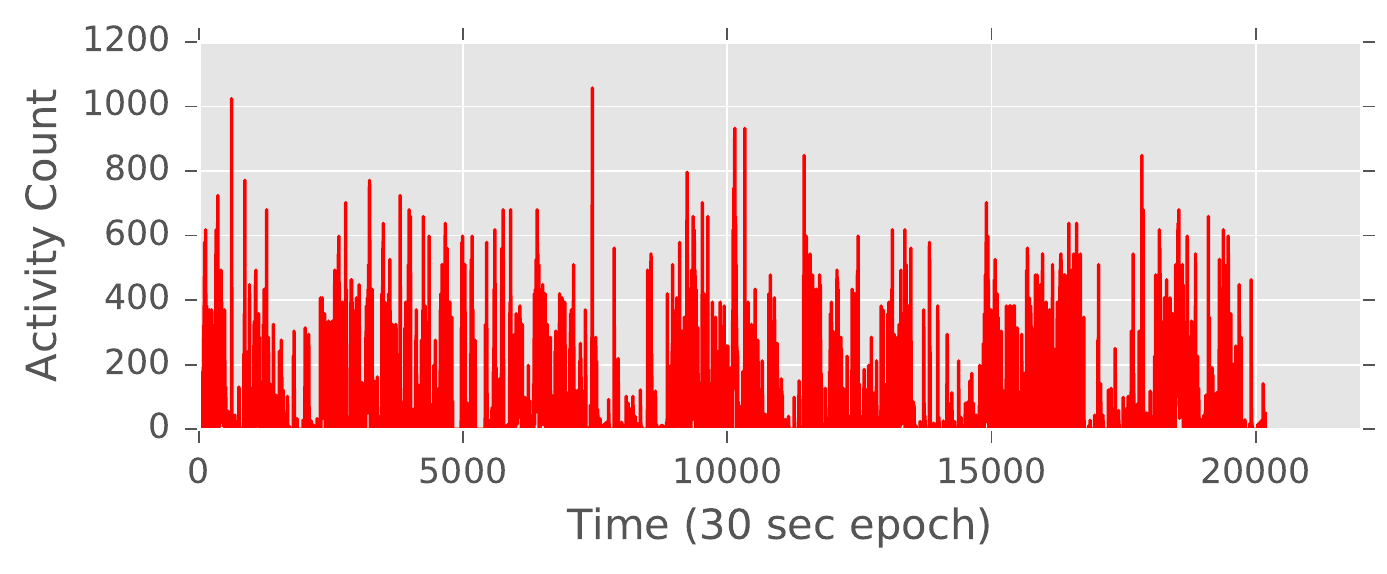}
      \vspace{-1em}
  \caption{Activity time-series for a subject over a week.}
    \vspace{-1em}
  \label{fig:activity}
\end{figure}

\begin{noindlist}
\item \textbf{Human activity for health informatics:} Wearable sensors have mostly been used for human activity recognition (HAR) task in machine learning~\cite{bulling2014tutorial,alsheikh2016deep}, while medical practitioners perform a manual examination on the actigraphy data for diagnosing mostly sleep-disorders~\cite{sadeh2011role}. 
Recent works have tried using actigraphy data for quantifying sleep quality using deep learning~\cite{sathyanarayana2016impact} or for actively monitoring human behavioral patterns~\cite{althoff2017harnessing}.  The novelty of our method is that we propose task-agnostic models rather than plain supervised learning.

\item \textbf{Representation learning:}
~\citeauthor{bengio2013representation} provide an overview of representation learning, used to construct a space that is discriminative for downstream tasks. 
It is based on ideas of better network convergence by adding (unsupervised) pre-trained vectors that encode the mutual information between the input features~\cite{goodfellow2016}. Recently, the area has made enormous progress in NLP, vision, and speech~\cite{collobert2011natural,hinton2012deep}. Of particular interest are the distributed bag-of-words (DBOW) architectures~\cite{mikolov2013distributed,Grover.Leskovec:16,saha2017regularized} optimized to predict the context of the structure, unlike continuous-bag-of-words (CBOW) that predicts the structure from its context. In a similar fashion, we use 
DBOW to capture local patterns in a time segment. Our novelty lies in integrating it in an adversarial setting~\cite{Ganin:2016} with an unsupervised predictor, which is a combination of DBOW with global context and ordinal constraints for activity time-series.

\item \textbf{Time series analysis literature:} 
These methods mainly use pair-wise similarity concept to perform 
classification~\cite{bagnall2017great} and clustering tasks, with a distance metric like Euclidean.  
Dynamic Time Warping~\cite{berndt1994using} is a widely used technique for finding similarity between two time-series which is computationally expensive due to its pair-wise similarity approach. This has led to the creation of symbolic representation techniques like SAX~\cite{lin2007experiencing} and BOSS~\cite{schafer2015boss}, that convert time-series into a symbolic sequence based on amplitudes or frequency analysis, respectively. SAX-VSM and BOSS-VS~\cite{schafer2016scalable} use tf-idf (term frequency-inverse 
document frequency) transformation of these symbolic sequences to get vector representation of sequence windows. 
HCTSA~\cite{fulcher2014highly} is an \emph{unsupervised}
time-series feature extraction engine with over 7800 feature space based on frequency-domain and time-domain analysis of the time-series, unlike the above supervised vector space models. These time-series models, however, cannot complement the supervised learning model, unlike our model's embeddings that can be used to initialize the architectures with back-propagation like neural networks. 
\end{noindlist}


\section{Our Approach} \label{sec:model}
In this section, we describe our method \acttovec. We first describe challenges, followed by the model components that address these challenges. 









\subsection{Challenges for \acttovec}
\label{subsec:acttovec}
To create a representational schema for activity time-series, the first natural challenge is determining the \textbf{right granularity} of the analysis unit. For example, consider the sample time-series in Figure \ref{fig:activity}, where the x-axis represents the time in 30 seconds epochs and the y-axis represents the activity levels (or counts), which in our setting are discrete values, ranging from $0$ to $5000$. 
Learning vector representations for each activity level might result in sparse vectors that are too fine-grained to be effective in the downstream tasks. Similarly, learning a representation with too big an analysis unit (\eg\ spanning a week) could result in generic vectors lacking required discriminative power. 
As we demonstrate later in our experiments, the right level of granularity is somewhere in between (a day span).  Within the analysis units, the \textbf{relative magnitude} of the signal values (\eg\ `10' $<$ `15') should also be taken into account. 

Considering granularity of analysis unit shorter than the sequence length posits another challenge -- how to capture the \textbf{global contextual dependencies} between the units. Since the units are parts of a sequence that describes a person's activity over a timespan, they are likely to be inter-dependent which should be captured in the representations. 
The same activity can look very different across the subjects owing to \textbf{subject-specific} noise and environment,  \eg\ how they wear the device on their wrist. Our \acttovec\ model addresses these challenges as described in the next section. 




\subsection{Problem Formulation and Time Granularity}
\begin{table}[b!]
    \caption{Notation table }
    \label{table:notation}
    \centering
  \resizebox{\columnwidth}{!}{
        \begin{tabular}{|l|L{8.5cm}|}
            \hline
            Notation & Description\\\hline
            $P$ & Set of subjects \\
            $S$ & Time-series dataset  \\
            $S_n$ & $n$-th time-series sequence $(t_1,\dots,t_l)$ from set $S$ \\
            $T_k$ & Time-series segment $(t_a,\dots,t_{a+L-1})$  \\
            $\mathcal{T}$ & Set of time-segments\\
            $g$ & Time-granularity of the model \\
            $d$ & Dimension of distributed representation space\\
            $\mathbf{\Phi}$ & Embedding Matrix representing the mapping function\\
            $\mathbf{\Phi}(T_k)$ & Representation of time-segment $T_k$\\
            $t_j$ & A time-series value/symbol in the segment $T_k=(t_a,\dots,t_{a+L-1})$\\
            $T_i$ & A neighboring segment of time-segment $T_k$ under consideration\\
            $\mathcal{N}(T_k)$ & Neighborhood of time-segment $T_k$\\
            $\Ls_s$ & Segment specific loss based on the pair $(T_k, t_j)$\\
            $\Ls_o$ & Ordinal Regression loss based on the pair $(T_k, t_j)$\\
            $\Ls_{nc}$ & Neighbor prediction loss based on the pair $(T_k, T_i)$ with $T_i \in \mathcal{N}(T_k)$\\
 $\Ls_r$ & Smoothing loss based on the pair $(T_k, T_i)$ with $T_i \in \mathcal{N}(T_k)$\\
$\Ls_a$ & Adversarial loss based on the pair $(T_k, p)$ with $p \in P$    \\\hline
        \end{tabular}
        }
\end{table}


\begin{figure*}[tp]
  \centering
  \fbox{
   \scalebox{0.18}{\includegraphics*[viewport=10 110 2075 1180]{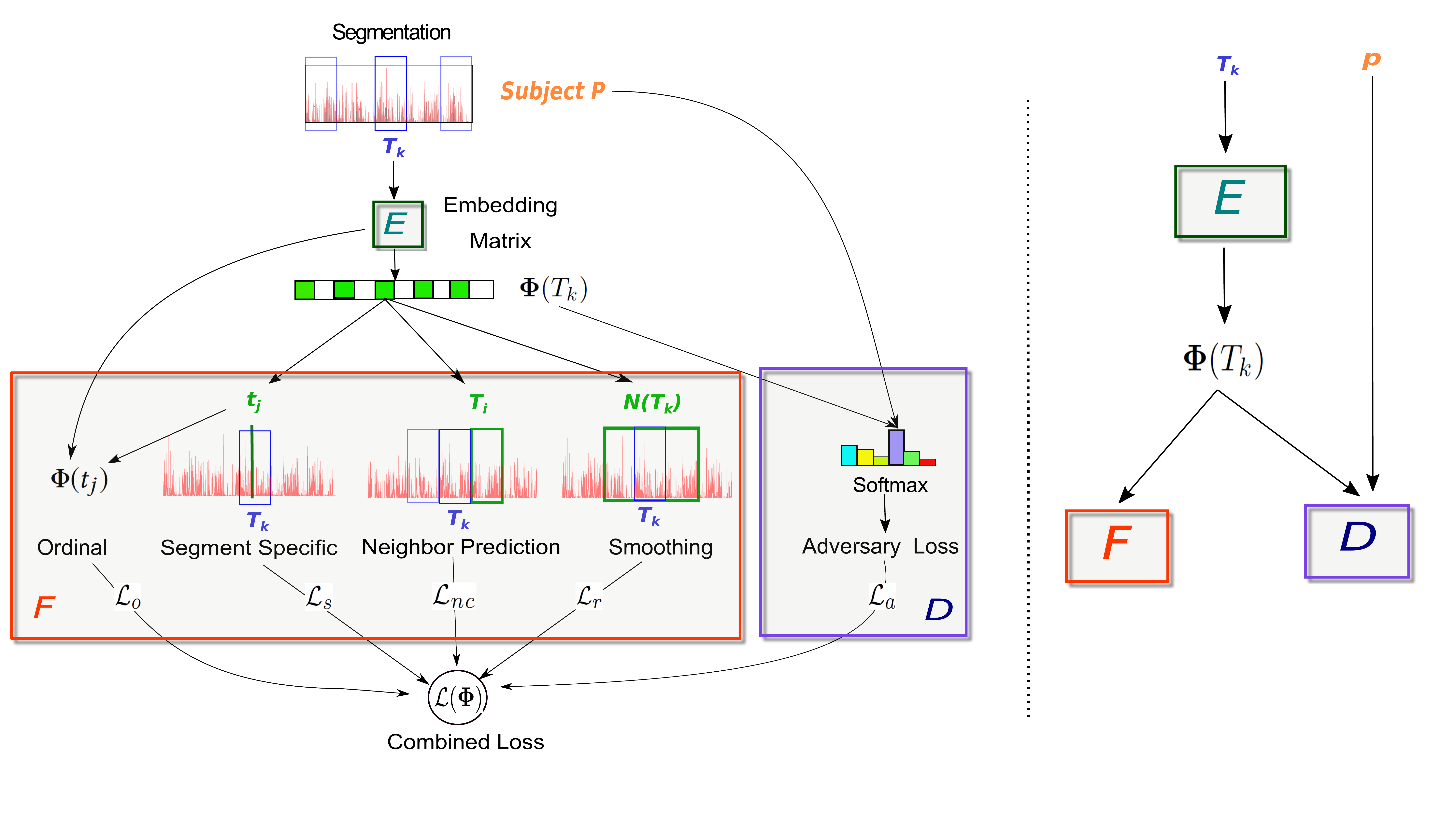}}
  }
  \caption{Graphical flow of \acttovec's components: encoder E (embedding matrix), regularized predictors F, and the subject discriminator D. Figure on left shows the sub-losses of the three components, while figure on right shows the overall schema of a three-player game between F, D, and E. First, segmentation is done based on the time granularity. For a selected segment $T_k$, its embedding $\ph(T_k)$ is first looked up from E. The embedding is then used by the predictors in F and the discriminator D. The encoder E plays a cooperative game with F to allow it to induce the necessary information. E also plays an adversarial min-max game against D to prevent it from identifying the subject from the embedding to promote subject invariance.} 
\label{fig:flowdiag}
\end{figure*}

Let $S = \{S_1, S_2, \cdots, S_N\}$ denote a set of activity sequences for $P$ unique subjects, where each sequence $S_n = (t_1,t_2,\dots,t_l)$  is $l$-length activity (\eg\ step counts) for a subject $p$ over a time period (a week here). Let $g \in \{$30 sec, 1 hour, 1 day, 1 week$\}$ specify the granularity we want to encode. We first break each sequence $S_n$ into $K$ consecutive time segments of equal length based on $g$ (top of Figure \ref{fig:flowdiag}). 
Let $T_k = (t_a,t_{a+1},\dots,t_{a+L-1}) \in \mathcal{T}$ be such a segment of length $L$ that starts at time $a$. Our aim is to learn a mapping function $\ph: \mathcal{T} \rightarrow \real^{d}$ to represent each time segment $T_k$ by a $d$ dimensional representation. Equivalently, if we represent each time segment in the dataset with a unique identifier (ID), the mapping function $\ph$ is a lookup operation on an embedding matrix of a single hidden layer neural network with non-linear activations. Our goal is to learn the embedding matrix by considering the segment's \emph{content} and \emph{context} while promoting subjective invariance. $S_n$ can be obtained by concatenating (or averaging) the $K$ segment-level vectors; we use concatenation. In this work, we consider the following time spans for a comparative analysis:
\begin{noindlist}
\item \textbf{Sample (\sampletovec):} representation for each 30-second sample. Our 20,160 length time-series yields a representation space of $\real^{20160 \times d}$. 
\item \textbf{Hour (\hourtovec):} representation for one-hour chunks, producing a vector space of $\real^{168 \times d}$. 
\item \textbf{Day (\daytovec):} embeds time-series at the level of a day span, giving us a   representational space of $\real^{7 \times d}$. 
\item \textbf{Week (\weektovec):} provides embeddings at the scale of a week, yielding a vector in $\real^{d}$ space.  
\end{noindlist}
For a given granularity level, \acttovec\ learns the mapping function $\ph$. 
{While it is possible to use a pre-processing step with change-points detection (CPD) to get the time-segments instead of manually setting the granularity, this step needs careful adjustments to set the thresholds for CPD models adding to further complexity of the model. In principle, we can have a space search over the possible granularities instead of using the pre-defined set above. We skip that step in this work for the sake of simplicity. Instead, we demonstrate the model's behavior for the choice of granularities that are intuitive to humans.} Figure~\ref{fig:flowdiag} presents the graphical flow of \acttovec. 

\noindent Our model relates to the sequential methods like LSTM by taking into account the global temporal dependencies.  Considering the inter-segment and intra-segment context is analogous to a shallow auto-encoder with very dense localized connections and sparse neighboring connections. While we leverage the discrete valued nature of our time-series, we can apply this method to continuous valued time-series by discretization, as done in the traditional time-series literature. The adversarial setting motivated by the  environmental noise, which might not apply to other domains. We first present the loss components, and then the combined loss. 


%



\subsection{Modeling Segment Content in \acttovec} \label{subsec:content}

We use two loss functions in \acttovec\ to capture the content of a segment -- the ordinal relation between time-series values and their co-occurrence patterns.  

\subsubsection{\textbf{Segment-Specific Loss}}
We use the segment-specific loss to learn a representation for each time segment by predicting its own values. 
Given an input sequence $T_k = (t_a,t_{a+1},\dots,t_{a+L-1})$, we first map it to a unique vector $\ph(T_k)$ by looking up the corresponding vector in the shared embedding matrix $\ph$. We then use $\ph(T_k)$ to predict each symbol $t_{j}$ sampled randomly from a window in $T_k$. However, using a softmax layer for the prediction is very computationally expensive. To compute the prediction loss in an efficient manner, we use Noise-Contranstive Estimation~\cite{gutmann2012noise} as an alternative to softmax: 
\begin{eqnarray}\label{eq:granloss}
\Ls_s(\ph,\mathbf{W_s}|T_k,t_j) = - \log\sigma(\mathbf{w}_{t_j}^\top \ph (T_k)) \\ \nonumber
 \hspace{-0.5em} - \hspace{-0.5em} \sum_{m=1}^{M} \mathbb{E}_{t_m \sim \nu(t)} \log\sigma(-\mathbf{w}_{t_m}^\top \ph (T_k))
\end{eqnarray}
\noindent where $\sigma$ is the sigmoid function defined as $\sigma (x) = 1 /(1+ e^{-x})$, $\mathbf{w}_{t_j}$ and $\mathbf{w}_{t_m}$ are the weight vectors associated with ${t_j}$ and ${t_m}$ symbols, respectively, $\nu(t)$ is the  noise distribution from which negative example ${t_m}$ is sampled, and $M$ is the number of negative examples sampled for each positive example. In our experiments, we use unigram distribution as the noise distribution with $M = 12$.

Since we ask the same segment-level vector to predict its symbols, the model captures the overall pattern of a segment. Note that except for \sampletovec, the model learns embeddings for both segments and symbols. 
A segment-based approach is commonly used in the time-series analysis, though among the representational models only models like SAX-VSM \cite{senin2013sax} look at the co-occurrence statistics, with a \textit{bag-of-words} assumption. 
 
\subsubsection{\textbf{Ordinal Regression Loss}} When predicting an activity symbol, the segment-specific loss above treats each symbol independently. However, since the symbols represent activity levels, there is a natural ordering in their values (\eg\ `5' $>$ `1'), which should be preserved in representations.  
In order to embed this ordinal relation, we use ordinal regression loss while learning the representation for each activity value:
\begin{equation}\label{eq:ordloss}
\begin{split}
\Ls_o(\ph,\theta,\mathbf{w}_o|t_j)  &= - \sum_{c=1}^{V} \mathbb{I}(t_j=c) \log \Big[\sigma(\mathbf{\theta_c}- \mathbf{w}_o^\top \Phi(c))\\ 
&\quad- \sigma(\mathbf{\theta_{c-1}}- \mathbf{w}_o^\top \Phi(c))\Big]
\end{split}
\end{equation}
\noindent where $\sigma(x)$ is the sigmoid function as defined before, $\mathbb{I}$ is the indicator function, and $V$ is the number of distinct discrete values that time-series can take (0-5000 in our case). Here, $\sigma(\mathbf{\theta_c}-\mathbf{w}_o^\top \Phi(c)) = p(t_j\leq c | \Phi(c))$ is the cumulative probability of $t_j$ being at most $c$ with $\theta_c$ being the ordered threshold for the regression, such that $\theta_j > \theta_i$, $\forall  j>i$.

\noindent \textbf{\emph{Remark:}} In principle, we can integrate the ordinal relation in Equation~\ref{eq:granloss} with the NCE loss. However, owing to the hierarchical nature of our algorithm, the segment ID is also predicted while learning the symbols and segments representations. Since these segment-level IDs do not have an ordinal relation with the symbol IDs, we resort to using an ordinal loss applicable only to time-series symbols. 

\subsection{Modeling Segment Context in \acttovec}


Loss functions presented above capture local patterns in a segment. However, since the segments are contiguous and describe activities of a person, they are likely to be related. For example, after a strenuous activity, there might be lighter activity periods. Likewise, one can expect a smooth transition from one segment to the next. The algorithm should capture relations between proximal segments. With this motivation, we use two loss functions to model this relationship.

\subsubsection{\textbf{Neighbor Context Loss}}
Similar to activity symbols, each segment in the dataset is assigned a unique identifier that we can use to look up its corresponding vector in the embedding matrix or to predict the segment ID. We first formulate the relation between neighboring segments by asking the current segment vector $\ph(T_k)$ (\ie\ looked-up vector for segment $T_k$) to predict its neighboring segments in the time-series: $T_{k-1}$ and $T_{k+1}$.  If $T_i$ is a neighbor to $T_k$, the neighbor context loss is the neighbor prediction task using NCE as before: 
\begin{eqnarray}\label{eq:conloss}
\Ls_{nc}(\ph,\mathbf{W_{nc}}|T_k,T_i) =  - \log\sigma(\mathbf{w}_{T_i}^\top \ph (T_k)) \\ \nonumber
 \hspace{-0.5em} - \hspace{-0.5em} \sum_{m=1}^{M} \mathbb{E}_{T_m\sim \nu(T)} \log\sigma(-\mathbf{w}_{T_m}^\top \ph (T_k))
\end{eqnarray}
\noindent where, $\mathbf{w}_{T_i}$ and $\mathbf{w}_{T_m}$ are the weight vectors associated with $T_i$ and $T_m$ segments, respectively. The noise distribution $\nu(T)$ is as described before over segment IDs.  





\subsubsection{\textbf{Smoothing Loss}}
While the previous objective attempts to capture neighborhood information, we also hypothesize that there is a ``continuity" between neighboring segments.  
The learning algorithm should discourage any abrupt changes in the representation of proximal segments. We apply \emph{smoothing between the neighboring segments} by minimizing the $l_2$-distance between the representations of the neighbors:
\small
\begin{equation}\label{eq:regloss}
\begin{split}
\Ls_r(\ph|T_k,\Ns(T_k)) &= \frac{\eta}{\mid \Ns(T_k) \mid} \sum_{T_c \in \Ns(T_k)} \norm{\ph(T_k) - \ph(T_c)}^2 
\end{split}
\end{equation}
\normalsize
\noindent where $\Ns(T_k)$ is the set of time-segments in proximity to $T_k$ and $\eta$ is the smoothing strength parameter. Note that the smoothing loss is not applicable to \weektovec.

\subsection{Modeling Subject Invariance in \acttovec}

One challenge in dealing with human activity data is that it heavily depends on the subject's environment. To promote subject invariance, we use an \emph{adversary} loss~\cite{Ganin:2016}. Let $P$ be the number of unique subjects. We use a multi-class classifier as a 
\emph{discriminator} to predict the {time-segment's source (subject) $s \in \{1,\dots,P\}$
from the encoded segment representation $\ph(T_k)$. In other words, the discriminator tries to identify the subject from whose activity time-series the \emph{encoder} (the embedding matrix) has derived the segment's representation. Note that we want to emphasize the invariance from the source subject of the time-segment and not from the particular sequence from which the segment is derived. Hence, if we have multiple sequences for the same subject, the sequences will share the same subject $s$ in our model.}  Formally, the discriminator is defined by a soft-max: 
\begin{eqnarray}
p(s=p|\Phi(T_k),\mathbf{U}) = \frac{\exp (\mathbf{u}_{p}^T \mathbf{\Phi(T_k)} )} {\sum_{p'} \exp (\mathbf{u}_{p'}^T \mathbf{\Phi(T_k)} )} \label{eq:softmax}
\end{eqnarray}
\noindent where $\mathbf{u}_{p}$ is the weight vector associated with subject $p$, and $\mathbf{U}$ defines all the parameters of the discriminator. We use a cross-entropy loss to optimize the subject discriminator:
\small
\begin{eqnarray}
\Ls_d(\mathbf{U}|\ph,T_k,s=p) = - \sum_{s=1}^{P} \mathbb{I} (s=p) \log p(s=p|\Phi(T_k),\mathbf{U}) 
\label{eq:domainloss}
\end{eqnarray}
\normalsize
\noindent where $p$ is a subject from whom the segment $T_k$ is encoded.  

With a goal to promote subject invariance, we put the encoder (the embedding matrix) in adversary with the discriminator. The encoder attempts to encode features that are indistinguishable to the subject discriminator by minimizing (negative of discriminator loss):
\small
\begin{eqnarray}
\Ls_a(\ph|\mathbf{U},T_k,s=p) =  \sum_{s=1}^{P} \mathbb{I} (s=p) \log p(s=p|\Phi(T_k),\mathbf{U}) \label{eq:domainloss2}
\end{eqnarray}
\normalsize
\subsection{Combined Loss for \acttovec\ }
We define our \acttovec\ model as the combination of the losses described in Equations~\ref{eq:granloss}, ~\ref{eq:ordloss},~\ref{eq:conloss}, ~\ref{eq:regloss}, and ~\ref{eq:domainloss2}:
 \small
\begin{equation}\label{eq:totalloss}
\begin{split}
\Ls(\ph) &= \sum_{n=1}^{N} \sum_{\substack{T_k \in S_n}} \hspace{-0.5em} \sum_{\substack{t_j \in T_k\\ T_i \in \Ns(T_k)}} \hspace{-0.6em} \Big[\underbrace{\Ls_s(\ph,\mathbf{W}|T_k,t_j) + \mathbf{\beta} \Ls_o(\ph,\theta,\mathbf{w}_o|t_j)}_{\text{\normalsize Segment Content\small}} + \\ &\quad \underbrace{\Ls_{nc}(\ph,\mathbf{W}|T_k,T_i)  + \Ls_r(\ph|T_k,\Ns(T_k))}_{\text{\normalsize Segment Context\small}} + \underbrace{\mathbf{\lambda}  \Ls_a(\ph|\mathbf{U},T_k,s)}_{\text{\normalsize Adversarial\small }}\Big]
\end{split}
\end{equation}
\normalsize
\noindent where $\beta > 0$ and $\lambda > 0$ are the relative weights for the ordinal regression loss and the subject invariance loss, respectively. Concurrently, we also minimize the discriminator loss in Equation \ref{eq:domainloss}. {As shown at the right in Figure \ref{fig:flowdiag}, the training of \acttovec\ involves an optimization game between three players: the encoder (E), the combined predictor (F), and the discriminator (D). E plays a cooperative game with F to allow it to induce the necessary information. E also plays an adversarial min-max game against D to prevent it from identifying the subject from the encoded vector to promote subject invariance.
Algorithm \ref{alg:jnt} presents our training procedure based on stochastic gradient descent (SGD). 

{The main challenge in adversarial training is to balance the two components -- the combined loss in Eq. \ref{eq:totalloss} vs. the discriminator loss in Eq. \ref{eq:domainloss}, as shown (right) in Figure \ref{fig:flowdiag}. If one player becomes smarter, its loss to the shared encoder (embedding matrix) overwhelms, and the training fails to converge \cite{ArjovskyCB17}. In our experiments, the discriminator converges much faster. To stabilize the training, we update the discriminator once every five gradient steps of the algorithm, chosen randomly.} Also, we follow the weighting schedule proposed by \cite[p. 21]{Ganin:2016}, that initializes $\lambda$ to $0$, and then changes it gradually to $1$ as training progresses.}  
{Through our experiments we demonstrate that the intuitions captured by the components are synergistic since they improve the performance incrementally as the components are added.}

\setlength{\textfloatsep}{0.1in}
\begin{algorithm}[t!]
\footnotesize
\SetAlgoLined
\KwIn{set of time-series  $S = \{S_1, S_2, \cdots, S_N\}$ with $S_p=(t_1,t_2,\dots,t_l)$, granularity level $g$}
\KwOut{learned time-series representation $\mathbf{\Phi}(S_n)$}
\SetKwBlock{Begin}{}{end}
 Segment time-series $S_p$ based on the granularity $g$;\\
 Initialize parameters: $\mathbf{\ph}$, $\mathbf{W_s}$, $\mathbf{W_{nc}}$, $\mathbf{U}$, $\mathbf{w_o}$, $\theta$; \\
 Compute noise distributions: $\nu(t)$ and $\nu(T)$ \\ 
 \Repeat {convergence}{
 Permute $S$;\\
\For{each time-series sequence $S_n \in S$}{
\For{each time-segment $T_k \in S_n$}{ 
    \hspace{-0.0cm}\For{each time-series sample $t_j \in T_k$}{ 
          \hspace{-0.3cm} - Take $(T_k, t_j)$ as a +ve pair and sample $M$ -ve pairs $\{({T_k}, t_m)\}_{m=1}^{M}$ from $\nu(t)$; \\
          \hspace{-0.3cm} - Perform updates, $\nabla \Ls_s$,  $\mathbf{\beta} \nabla \Ls_o$; \\
          \hspace{-0.3cm} - Sample a neighboring time-segment $T_i$ from sequence $S_n$; \\
          \hspace{-0.3cm} - Take $(T_k, T_i)$ as a +ve pair and sample $M$ -ve pairs $\{({T_k}, T_m)\}_{m=1}^{M}$ from $\nu(T)$; 
        
          \hspace{-0.3cm} - Perform updates, $\nabla\Ls_{nc}$, $\nabla\Ls_r$;\\
        \hspace{-0.3cm} - Perform update $\lambda \nabla \Ls_a$; 
    } 
    }
   }
}   
\caption{Training \acttovec\ with SGD}
\label{alg:jnt}
\end{algorithm}
%



\section{Experimental Settings} \label{sec:settings}

{In this section, we describe experimental settings --- datasets, the prediction tasks on which we evaluate the embeddings, the baselines models, and parameter selection.} 

\subsection{Datasets} 
We use Study of Latinos (SOL)~\cite{sorlie2010design} and  Multi-Ethnic Study of Atherosclerosis (MESA)~\cite{bild2002multi} 
datasets. SOL has physical activity and clinical data for 1887 subjects, while MESA only has activity data for 2237 subjects. This simulates the scenario where disorder condition labels are available only for a portion of users. These datasets contain activity data (actigraphy) from each subject for a minimum of 7 days measured with wrist-worn Philip's Actiwatch Spectrum. 
The time-series for each subject is sampled at a rate of 30 seconds. 
Actigraphy records the mean activity count per 30 seconds, providing us with a signal that can only take integer values. Mean activity count is reported in ZCM (Zero Crossing Mode) that counts the number of times the waveform crosses zero. This can be generalized to steps since they calculate the intensity. This makes embedding the input symbols straightforward, without a dedicated discretization step for our proposed \acttovec\ method. 

The datasets are not skewed with respect to the class distribution of different prediction tasks described in the next section. 
A very few missing values {($<1\%$)} observed in the dataset were replaced by unknown (UNK) token. Any future unseen or out-of-vocabulary (OOV) signal value can be handled by a procedure that assigns representation from averaging out neighboring ordinal signal values rather than a generic unknown symbol assignment. 



\subsection{Prediction Tasks} \label{subsec:predtasks}

We evaluate the effectiveness of the learned embeddings on the following health disorder prediction tasks:  

\begin{noindlist}
\item \textbf{Sleep Apnea:} Sleep apnea syndrome is a sleep disorder characterized by reduced respiration during the sleep time. We use the Apnea-Hypopnea Index (AHI) at 3\% desaturation level with AHI$<$5 being characterized as \emph{non-apneaic}, while  AHI $>5$ indicating a \emph{mild-to-severe-apnea}. 
\item \textbf{Diabetes:} Diabetes (type 2) is the body's insensitivity to insulin, leading to elevated levels of blood sugar. Task is defined as a three-class classification problem, to decide whether a subject is a \emph{non-diabetic}, \emph{pre-diabetic}, or \emph{diabetic}. 

\item \textbf{Hypertension:} Hypertension refers to abnormally high levels of blood pressure, an indicator of stress. Hypertension prediction characterizes a binary classification problem for increased blood pressure.

\item \textbf{Insomnia:} Insomnia is a sleep disorder characterized by an inability to fall asleep easily, leading to low energy levels during the day. We use a 3-class prediction problem for classifying subjects into \emph{non-}, \emph{pre-} and \emph{insomniac} groups. 
\end{noindlist}

\subsection{Baseline Models}

We compare our method with a number of naive baselines and existing systems that use time-series representations:


\noindent \textbf{(a) Majority Class:} This baseline always predicts the class that is most frequent in a dataset.

\noindent\textbf{(b) Random:} This baseline randomly picks a class label.

\noindent\textbf{(c) SAX VSM:} SAX-VSM ~\cite{senin2013sax} uses SAX, one of the most  widely used time-series representation technique.

\noindent\textbf{(d) BOSS:} BOSS~\cite{schafer2015boss} 
is a symbolic  representational learning technique that uses Discrete Fourier 
Transform (DFT) of time-series windows. 
BOSS creates equal sized bins from histograms of 
DFT coefficients, which are then assigned representational symbols. Labels are assigned based on the class that gets the highest similarity score using nearest neighbor approach.

\noindent\textbf{(e) BOSSVS:} BOSS in Vector Space~\cite{schafer2016scalable} is similar to SAX-VSM and creates vector space representation of the time-series from BOSS. BOSS is known to be one of the most accurate methods
on standard time-series classification tasks, with BOSS-VS performing marginally lower.

\noindent\textbf{(g) CNN:} We use a Convolutional Neural Network (CNN) with Conv-ReLU-AvgPool-BatchNorm-Dropout layers for supervised prediction. We add layers until we get no performance improvement on held-out set.


\noindent\textbf{(h) HCTSA:} Highly Comparative Time-Series Analysis~\cite{fulcher2014highly} is a time-series feature extraction engine with over 7800 features extracted with frequency and time domain analysis like Kurtosis. Unlike the time-series baselines above, this is an unsupervised method --- most relevant baseline to our method.

{\noindent\textbf{(i) LSTM:} We train an LSTM based unsupervised model similar to~\cite{sundermeyer2012lstm} that learns to encode sequences by predicting next time-series value (as a language model in NLP). Next, we use this pre-trained network to initialize the LSTM network that is further trained on the supervised learning tasks.

\subsubsection{Variants of \acttovec}
\noindent\textbf{(a) Unregularized models:} This group of models contain only two NCE loss components from Equation \ref{eq:totalloss}: $\Ls_s$ and $\Ls_{nc}$. In the Results section, we refer to these models as \sampletovec, \hourtovec, \daytovec, and \weektovec. 

\noindent\textbf{(b) Regularized models:} We add smoothing loss $\Ls_r$ to models in (a). This group includes \hourtovecreg~and \daytovecreg. We omit \sampletovecreg~since it performed extremely poorly on all the tasks. Recall that smoothing is not applicable to \weektovec. 

\noindent\textbf{(c) Ordinal loss model:} We use ordinal loss $\Ls_o$ with these models. We add this loss to \daytovecreg~model in (b), our best performing model as discussed in the next section. We omit other time-unit models for brevity.

\noindent\textbf{(d) Adversarial model:} These models use the \emph{adversary} loss $\Ls_a$ for our \daytovecreg~with ordinal loss.






\begin{table*}[t]
  \renewcommand{\arraystretch}{1.1}
  \caption{ $\mathbf{F_1}$ and \textbf{Speed} relative to \textbf{wall clock time} of our \texttt{\daytovecreg+O}+\texttt{A}. O refers to \textbf{O}rdinal and A to \textbf{A}dversarial.}
  \label{ApneaResults}
  \centering
  \resizebox{1.5\columnwidth}{!}{%
  \begin{tabular}
{|c|l|l|c|cc|cc|c|c|}
    \cmidrule[\heavyrulewidth]{2-10}
     \multicolumn{1}{c|}{} & Method & Clf. & Sleep Apnea & \multicolumn{2}{c|}{Diabetes} & \multicolumn{2}{c|}{Insomnia} & Hypertension & Speed\\
    \cmidrule{2-10}
    \multicolumn{1}{c|}{} & &  & $F_{1}$ & $F_{1}$-macro & $F_{1}$-micro & $F_{1}$-macro & $F_{1}$-micro & $F_{1}$ & \\
    \midrule
    \parbox[t]{0.25cm}{\multirow{6}{*}{\rotatebox[origin=c]{90}{\textbf{Supervised}}}}  & \texttt{Majority} & 0-R & 00.0 & 21.7 & 31.9 & 47.4 & 25.5 & 00.0 & -\\
    & \texttt{Random} & & 33.9 & 34.3 & 31.3 & 36.2 & 30.0 & 33.4 & - \\
    
\cline{2-10}
     & \texttt{SAX-VSM} &  & 00.0 & 38.6 & 24.3 & 47.4 & 25.5 & 00.0&-\\   
   & \texttt{BOSS} &  & 17.6 & 38.9 & 31.5 & 49.8 & 34.9 & 29.6&-\\  
   & \texttt{BOSSVS} &  & 11.7 & 40.1 & 32.7 & 47.5 & 33.1 & 31.3&-\\
   & \texttt{Task-specific} & {CNN} &  41.5 & 45.2 & 41.0 & 50.7 & 40.1 & 36.6 & 2.0x \\
    \midrule
     \parbox[t]{0.25cm}{\multirow{10}{*}{\rotatebox[origin=c]{90}{\textbf{Unsupervised}}}} 
     & \sampletovec & LR &  36.7 & 40.0 & 36.7 & 42.4 & 35.3 & 39.4& -\\ 
    &\hourtovec & LR  &  30.0 & 41.4 & 33.3 & 44.6 & 28.5 & 24.4& -\\ 
    &\hourtovecreg & LR  & 20.5 & 42.1 & 32.0 & 43.5 & 28.7&23.1& -\\ 
    &\daytovec & {LR} & 36.8 & 40.9 & 38.0 & 45.2 & 35.8& 40.3& 0.3x\\
    &\daytovecreg & {LR} & \textbf{38.9} & 41.8 & \textbf{39.5} & 46.6 & \textbf{39.7} &\textbf{43.4} & 0.3x\\
    &\weektovec & {LR} & 14.5  & 40.6 & 34.1 & 44.2 & 31.5 & 18.7& -\\ 
\cline{2-10}
    
          
    &  \texttt{HCTSA} & LR &  20.3 & 40.0 & 35.0 & 46.7 & 33.7 & 22.0 & 8.2x\\
     & \texttt{LSTM} & LSTM & 32.2 & 41.4 & 33.3 & 46.1 & 30.4 & 37.8 & 10.5x \\
     & \pbox{20cm}{\texttt{\daytovecreg+O}} & {LR}  & 40.5 & 45.3 & 40.2 & 50.9 & 40.3 & 43.6 & 0.4x\\
     & \pbox{20cm}{\texttt{\daytovecreg+O}+\texttt{A}} & {LR}  &  \blue{\textbf{43.6}} & 45.8 & \blue{\textbf{42.5}} & 55.7 & \blue{\textbf{41.4}} & \blue{\textbf{44.1}} & \textbf{1.0x}\\\bottomrule
\end{tabular}
}

\end{table*}

\vspace{-0.25em}
\subsection{Hyper-Parameter Selection}
We use 80\%,10\%,10\% split for train, validation, and test sets repeated 10 times, and we report the mean scores. As mentioned earlier, we only have the disorder task labels for the SOL dataset. For the supervised models we only use the SOL dataset, while for the unsupervised models we use the combined SOL and MESA data. The embedding size of $d$=$100$ was fixed for all the models. The weighting parameters $\lambda$ and $\beta$ were chosen to be 0.05 and 0.5, respectively. The remaining hyper-parameters in \acttovec\ are: window size ($w$) for segment-specific  loss, number of neighboring segments ($|\Ns(T_k)|$) and regularization strength ($\eta$) for \daytovec\ and \hourtovec.  We tuned for $w \in \{12,20,30,50,100,120,500\}$, $\eta \in \{0,0.25,0.5,0.75,1\}$, and $|\Ns(T_k)| \in \{2,4\}$ on the development set. We chose $w$ of size 20, 20, 30, and 50 for \sampletovec, \hourtovec, \daytovec, and \weektovec, respectively. The $\eta$ of 0.25 and 0.5 were chosen for \daytovec\ and \hourtovec, respectively. The neighbor set size of 2 was chosen. For the CNN baseline, 3, 4, 3, and 3-layered network were used for sleep-apnea, diabetes, insomnia, and hypertension, with a dropout of 0.5 trained with Adam Optimizer. We tuned all the parameters for maximizing the $F_1$ scores.



\section{Results and Discussion} \label{sec:results}

In this section, we present our results for the four prediction tasks described in the previous section. The results are presented in Table~\ref{ApneaResults} in four groups: \Ni baselines, \Nii existing time-series representation methods and CNN, \Niii our unregularized and regularized \acttovec\ variants, and \Niv our ordinal and adversarial \acttovec\ models. We show classification performance in terms of $\mathbf{F}_1$ scores.





\begin{noindlist}

\item \textbf{\emph{Performance on disorder classification tasks:}}
Since our goal is to evaluate the effectiveness of the learned vectors, we use simple \textbf{linear classifier} Logistic Regression (LR) with our \acttovec\ models. For the multi-class classification problems like Diabetes and Insomnia, we use One-vs-All classifiers, tuning for micro-$F_{1}$ score. We run each experiment 10 times and take the average of the evaluation measures to avoid any randomness in results. 
We can notice that \daytovec\ consistently gives absolute 2-4\%  improvements over the other \acttovec\ variants, and 6-20\% over the baseline time-series representation models and LSTM on $F_1$ scores on all tasks. The adversarial-ordinal-regularized variant of \daytovec\ gives the best results among all the variants. Adversarial variant performs at par or better than task-specific supervised CNN on all the tasks. Due to inherent noisy nature of activity time-series and long sequence length, the LSTM-based model did not work well.

\item \textbf{\emph{Selecting time granularity:}} Across the tasks, the adversarial ordinal \daytovecreg~outperforms 
all the models. Models with day level granularity perform better than all the other granularities as well as baseline methods. Among the \acttovec\ variants, the \weektovec~models perform the worst, while \hourtovec\ models perform just a bit better on an average. Hour and week-level models perform similarly to the baseline time-series methods. The high-dimensional 
models based on samples (\sampletovec) perform better than hour-level, week-level, 
and baselines. \daytovec\ produces marginally better results than the \sampletovec\ despite much lower dimensional space (2880x).  The level of granularity makes a lot of difference in the performance of our models. From the above results, we can conclude that 
while the low granularity level (\sampletovec) suffered from coarse
embeddings, the high granularity 
(\weektovec) level embeddings lost the ability to discriminate.

\begin{figure*}[t!]
\centering
\begin{subfigure}{.245\textwidth}
  \centering
  \includegraphics[width=\linewidth]{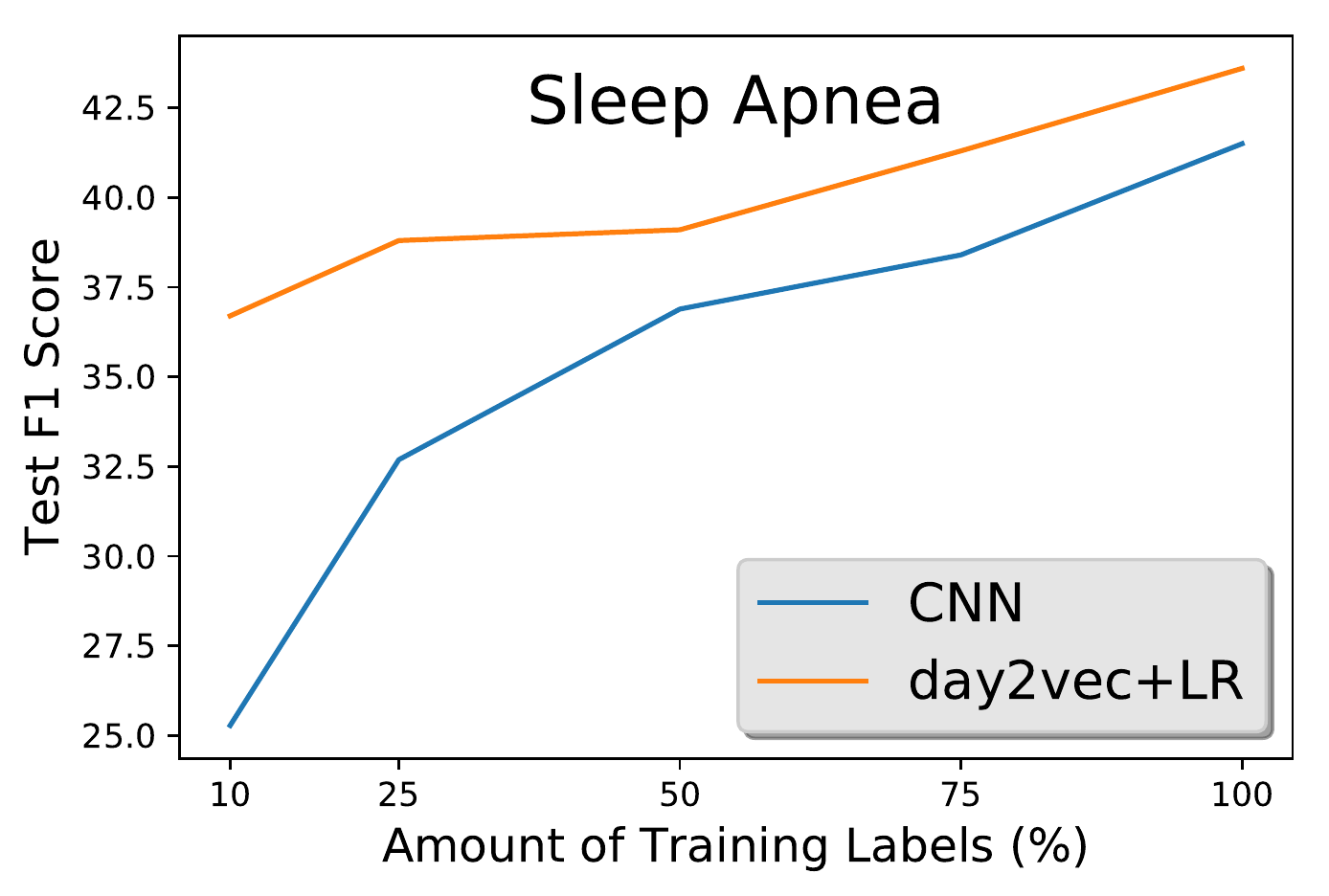}
\end{subfigure}%
\begin{subfigure}{.245\textwidth}
  \centering
  \includegraphics[width=\linewidth]{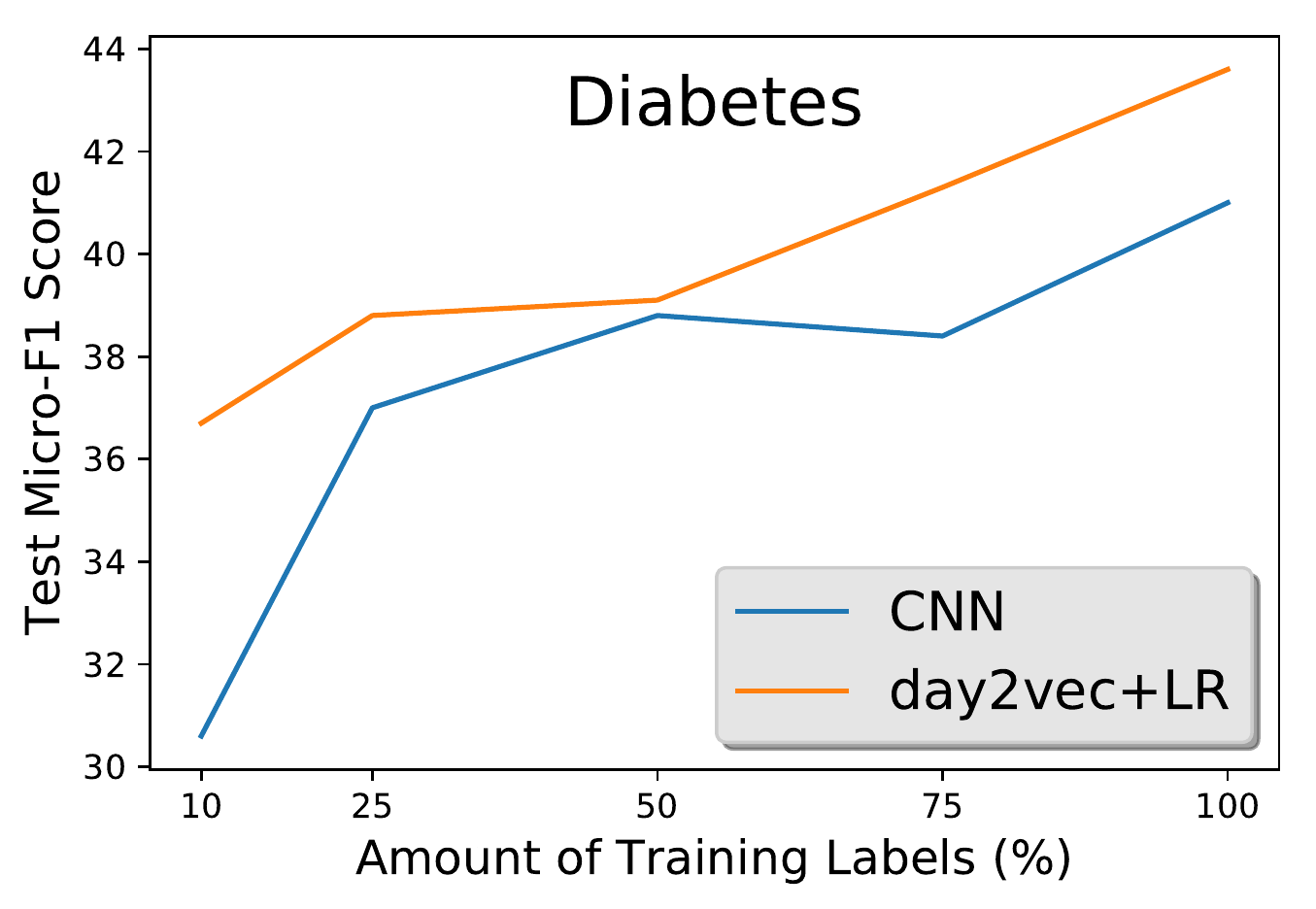}
\end{subfigure}
\begin{subfigure}{.245\textwidth}
\centering
  \includegraphics[width=\linewidth]{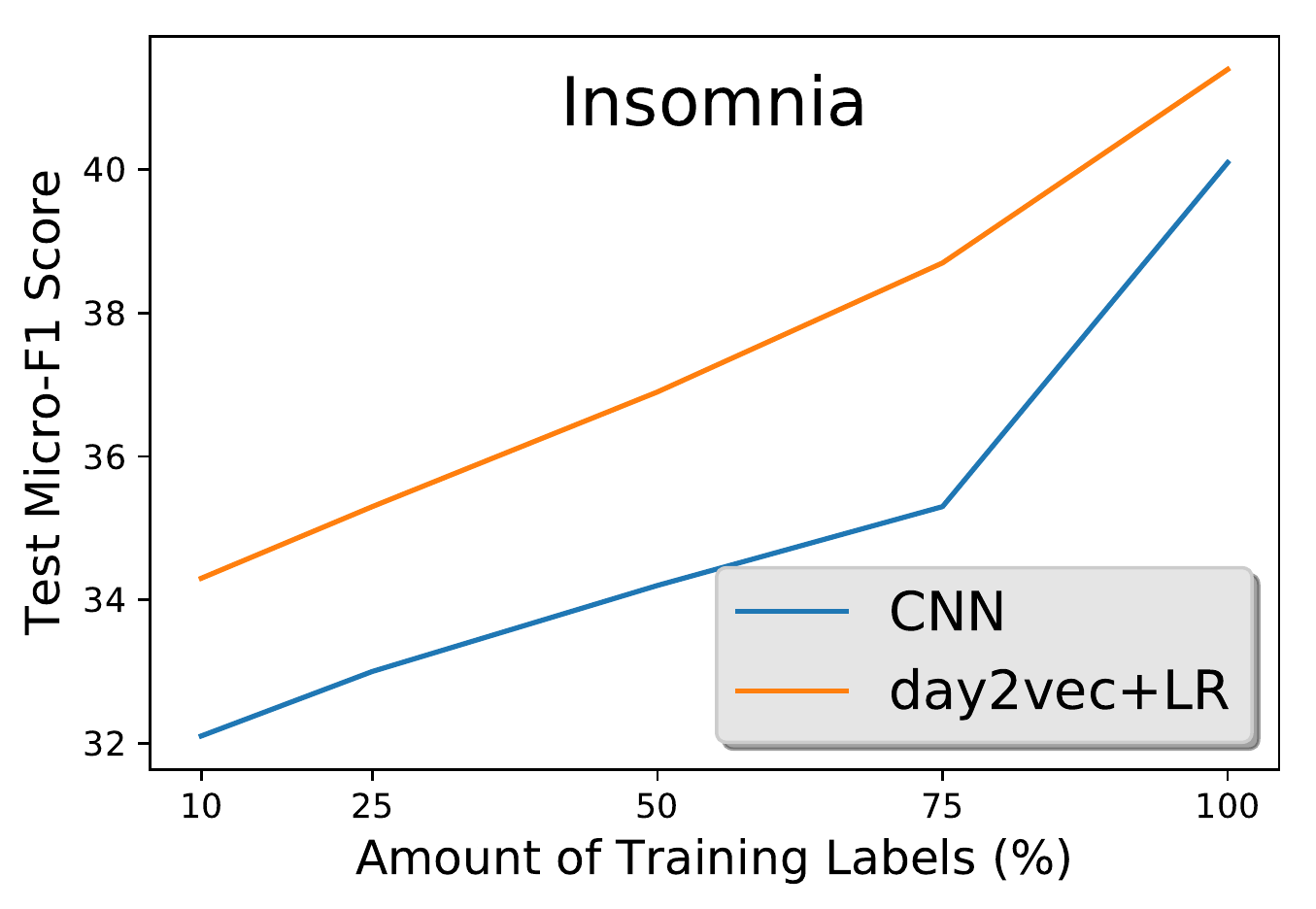}
\end{subfigure}
\begin{subfigure}{.245\textwidth}
\centering
  \includegraphics[width=\linewidth]{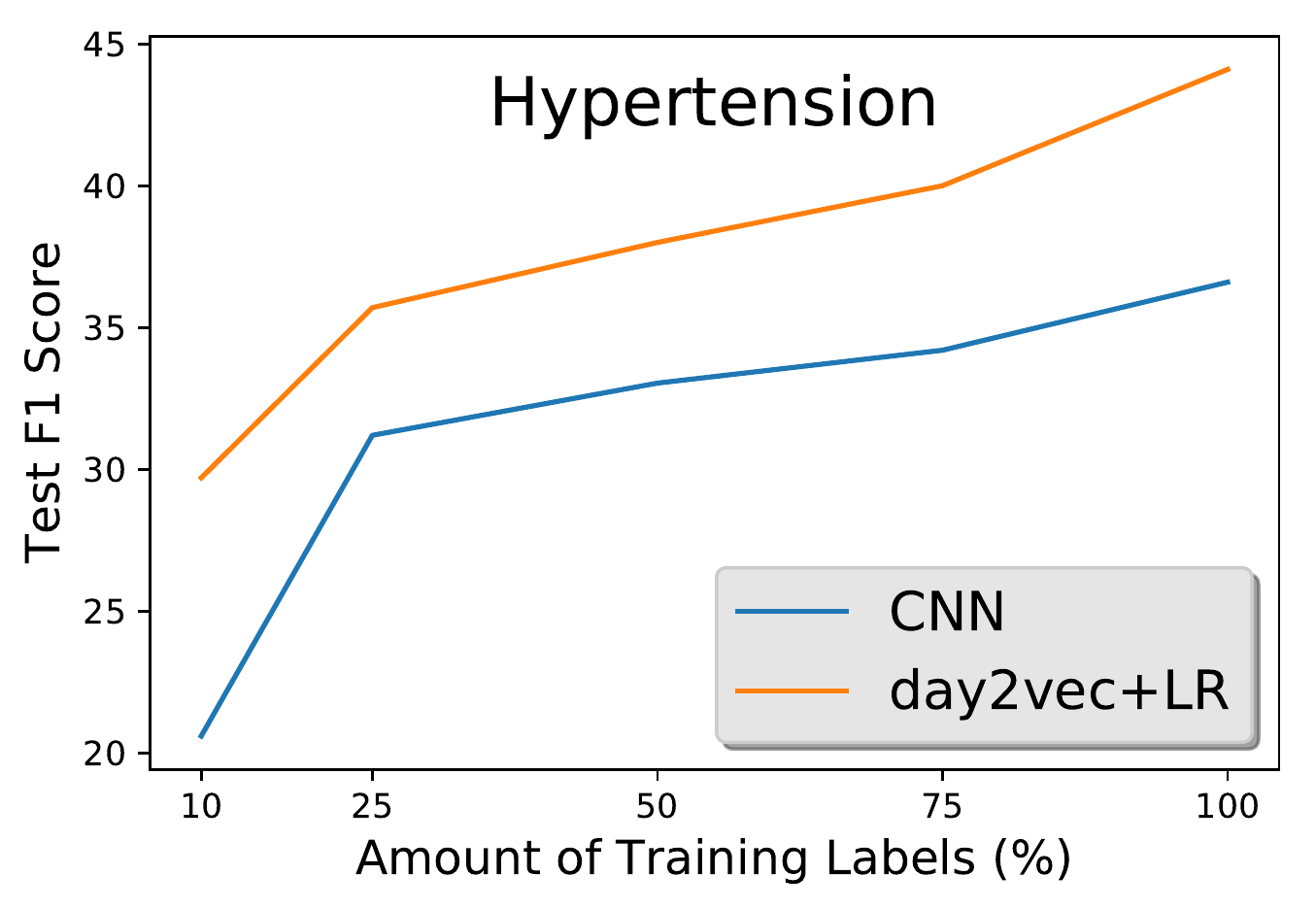}
\end{subfigure}
\caption{Comparison of our unsupervised \texttt{\daytovecreg+O}+\texttt{A} with LR vs. supervised CNN as a function of labeled data.}
\label{figure:UnVsS}
\end{figure*}

\begin{figure}[b!]
  \centering
  \includegraphics[width=0.45\textwidth]{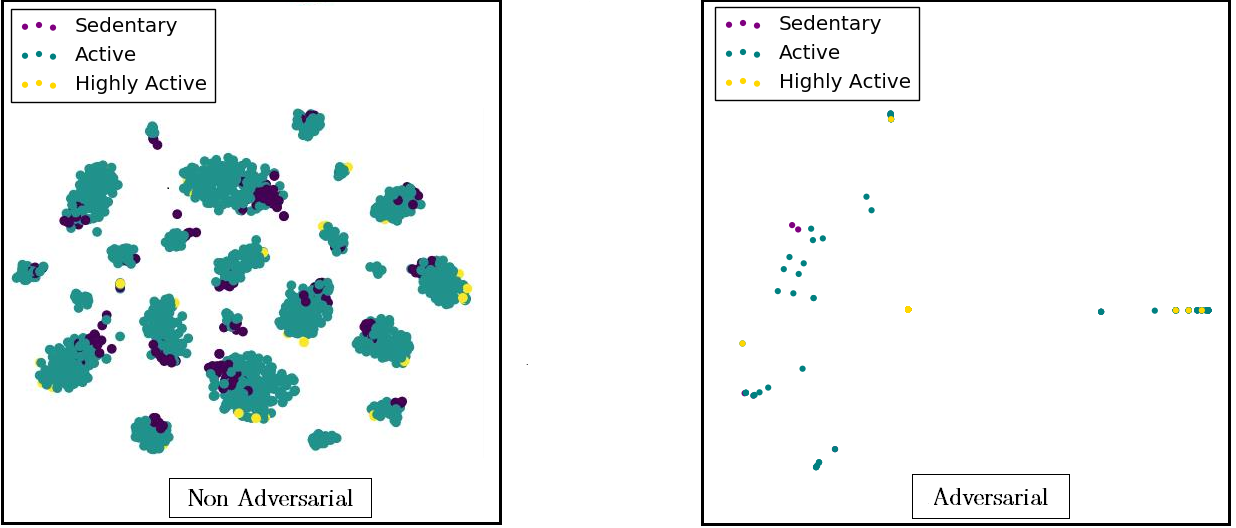}
  \caption{t-SNE visualization of subjects with regularized \daytovec\ on the left and adversarial \daytovec\ on the right, for all the subjects with respect to their level of activity. }
  \label{fig:tsne}
\end{figure}

\item \textbf{\emph{Effect of smoothing:}} 
{Intuition behind adding the temporal smoothing loss (Eq.~\ref{eq:regloss}) to our model was to test the hypothesis that human activities happen in continuity and follow a macro-routine. This should be reflected in neighboring time-segments making them similar in structure. 
As results suggest, the continuity hypothesis was misguided at the sample- and hour-level.  Regularization at the sample level made them lose the discriminative power for classification--- considering the noise in activity levels at such a fine granularity. 
{\hourtovecreg~exhibits a significant drop (sleep apnea) or at par performance compared to \hourtovec. Since humans tend to switch between different activity types at the order of hours or less, the hypothesis of continuity was inappropriate at hour level as well.} 
However, adding smoothing helps produce gains for \daytovec, our best model. We argue that smoothing regularization helps capture a higher level global context since humans structure  \emph{routines} at the level of the day, while we switch between activities on the order of hours or lower.} This is \textbf{supported by the periodogram analysis}, where day level frequencies dominate.

\item  \textbf{\emph{Effect of ordinal loss:}} Addition of ordinal loss improves the accuracy of our models, albeit marginally. While the other loss functions work on co-occurrence of activity values at local and global contexts, ordinal loss explicitly models the relationship between the magnitude of the activity values, \eg `25' $>$ `5'. While it can be argued that similar ordinal values would co-occur, hence reflected in the $\Ls_s$, adding an explicit ordinal constraint helps, though marginally.

\item \textbf{\emph{Effect of adversarial training:}} 
Adding the adversarial loss to the ordinal-regularized models improves the  $F_1$ scores across the tasks by 0.5\%-3.1\% in absolute numbers and 2\%-7.5\% in relative terms. The difference in performance is more than the 95\% confidence intervals around the repeated experimental means reported here. It can thus be concluded that making the representations invariant to the source (subjects) helps in removing the noise introduced by the subjects and their environments, owing to how they wear their device and their routines, making the embeddings more generalized. With an adversarial setting, we can create a representation space more relevant to the health condition of the subjects by removing the subject source domain.

Figure~\ref{fig:tsne} shows the t-SNE~\cite{maaten2008visualizing} plot of regularized \daytovec~(left) vs adversarial \daytovec~embeddings (right) for SOL dataset subjects. For each subject, we concatenate the embeddings from constituent day level time-segments to get the representation. We plot the lifestyles of the subjects as determined by the study questionnaire, identifying each subject as \emph{highly-active}, \emph{moderately-active}, and \emph{sedentary} person. We can notice that we get clusters or subject \emph{phenotypes} with nice separation within the cluster for each of the lifestyle type with non-adversarial \daytovec. Unsurprisingly, the clusters get very compact in the adversarial setting. The lifestyle classes get clustered markedly separately rather than forming in-cluster separations, observed in the non-adversarial setting. Hence, the adversary helps \acttovec\ with removing the subject-wide variance in the learned embeddings, while still capturing the properties of subject phenotypes. Also, adversary loss improved results on the disorder prediction tasks. Hence, by reducing subject level variance, the adversary loss helps encode a better global representation.

\item \textbf{Scalability:} As reflected in Table~\ref{ApneaResults}, our model takes much less training time compared to the deep unsupervised model like LSTM. In fact, our model takes much less time compared to the supervised CNN that only runs on the labeled data.
Since our model has only one hidden layer (\ie\ embedding layer) and uses NCE for training, it is \textbf{scalable} in practical settings compared to the deep neural models. Additionally, our model offers more \textbf{flexibility} to incorporate the alternative intuitions including human knowledge (\eg\ the day/hour level representational hierarchy) that would be difficult to do with other deep methods. \textit{Clearly, using more sophisticated deep neural models like deep auto-encoders or variational auto-encoders that are susceptible to noisy data like ours for a semi-supervised setting would pose scalability challenges.}


\textbf{\emph{Note:}} We restrict our models to a \emph{transductive} setting, where representations of test segments are learned along with training segments. However, it can be easily applied to a \emph{inductive} setting, where representations for unseen segments can be derived by a single backprop step~\cite{le2014distributed}.   
 
 \item \textbf{Supervised vs Unsupervised:} {To show the utility of unsupervised schema, we demonstrate the performance of supervised CNN vs. our adversarial \daytovec\ method as a function of the percentage of labeled data in Figure~\ref{figure:UnVsS}. Clearly, our model outperforms CNN across the board. The gap is drastic when the proportion of labeled data is \textbf{low}, which is usually the case in practice.}


\end{noindlist}

\section{Conclusions} \label{sec:conclusion}
In this work, we present a novel unsupervised representational learning technique, \acttovec\ that encodes human activity time-series by modeling local and global activity patterns. 
We train our model on two datasets and test on prediction tasks (four commonly occurring disorders). We find that day-level granularity preserves the best representations, which is not surprising since a day is a natural timescale for a full cycle of human activities. Our task-agnostic representational learning model using simple linear classifiers beats baseline time-series representation models on all the disorder prediction tasks. It even performs at par or better than supervised convolutional neural network baseline. Our model learns the representational features using a combination of non-linear loss functions, giving better performance on multiple tasks using simple linear classifiers. We further demonstrate that using adversarial loss along with our embedding encoder model helps increase the performance and generalizability of the embeddings. Additionally, our method is capable of complementing the supervised learning by initialization, unlike existing representation approaches. 

\balance

\bibliographystyle{aaai}
 {\bibliography{references}}
\end{document}